\documentclass[letterpaper]{article} 
\usepackage[draft]{aaai2026}  
\usepackage{times}  
\usepackage{helvet}  
\usepackage{courier}  
\usepackage[hyphens]{url}  
\usepackage{graphicx} 
\usepackage{natbib}  
\usepackage{caption} 
\usepackage{algorithm}
\usepackage{algorithmic}

\usepackage{multirow}
\usepackage{booktabs}
\usepackage{amsmath}
\usepackage{graphicx}
\usepackage{subcaption}
\numberwithin{equation}{section}

\usepackage{newfloat}
\usepackage{listings}
\DeclareCaptionStyle{ruled}{labelfont=normalfont,labelsep=colon,strut=off} 
\floatstyle{ruled}
\newfloat{listing}{tb}{lst}{}
\floatname{listing}{Listing}
\title{Optimal-Agent-Selection: State-Aware Routing Framework for Efficient Multi-Agent Collaboration}
\author{
    Jingbo Wang\textsuperscript{1}, Sendong Zhao\textsuperscript{1}\thanks{Corresponding author}, Haochun Wang\textsuperscript{1}, Yuzheng Fan\textsuperscript{1}, Ting Liu\textsuperscript{1}
}
\affiliations{

    \textsuperscript{1}Research Center for Social Computing and Information Retrieval,Harbin Institute of Technology, China
    \\
    \{jingbowang,sdzhao\}@ir.hit.edu.cn
}

\begin{document}

\maketitle

\let\svthefootnote\thefootnote
\let\thefootnote\relax
\footnotetext{This work has been submitted to the IEEE for possible publication. Copyright may be transferred without notice, after which this version may no longer be accessible.}
\let\thefootnote\svthefootnote

\begin{abstract}
The emergence of multi-agent systems powered by large language models (LLMs) has unlocked new frontiers in complex task-solving, enabling diverse agents to integrate unique expertise, collaborate flexibly, and address challenges unattainable for individual models. However, the full potential of such systems is hindered by rigid agent scheduling and inefficient coordination strategies that fail to adapt to evolving task requirements. In this paper, we propose STRMAC, a state-aware routing framework designed for efficient collaboration in multi-agent systems. Our method separately encodes interaction history and agent knowledge to power the router, which adaptively selects the most suitable single agent at each step for efficient and effective collaboration. Furthermore, we introduce a self-evolving data generation approach that accelerates the collection of high-quality execution paths for efficient system training. Experiments on challenging collaborative reasoning benchmarks demonstrate that our method achieves state-of-the-art performance, achieving up to 23.8\% improvement over baselines and reducing data collection overhead by up to 90.1\% compared to exhaustive search.
\end{abstract}

\section{Introduction}

\begin{figure}[t]
\centering
\includegraphics[width=0.9\columnwidth]{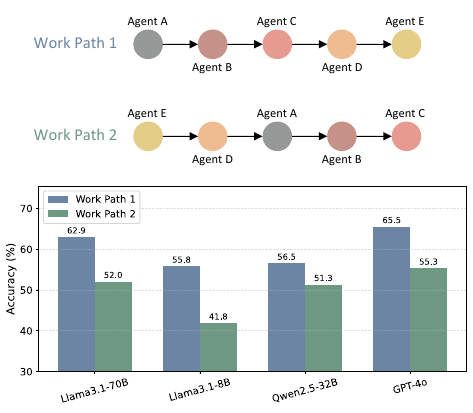} 
\caption{Accuracy comparison of two work paths for five agents (A-E) in the clinical prediction scenario. Work Path 1 (A→B→C→D→E) achieves significantly higher accuracy than Work Path 2 (E→D→A→B→C) across four large language models, indicating that agent order has a significant impact on distributed cooperative reasoning performance.}
\label{fig:path_comparison}
\end{figure}

The rapid advancement of large language models (LLMs) ~\cite{achiam2023gpt,touvron2023llama,liu2024deepseek}
has given rise to highly capable autonomous agents, enabling significant progress across a wide range of tasks~\cite{wang2023voyager,deng2023mind2web,hong2024data,qiao2024autoact}. Although single-agent systems built upon LLMs have demonstrated strong performance, their effectiveness can be limited when addressing complex scenarios that demand (1) multidisciplinary expertise or (2) collaborative problem-solving across domains. To overcome this challenge, 
recent research has witnessed a growing emphasis on multi-agent systems (MAS), where the coordination of multiple LLM-based agents has proven effective in improving code generation~\cite{huang2023agentcoder,qian2023chatdev}, reasoning~\cite{tang2023medagents,wang2024rethinking,zhang2025belle}, decision-making~\cite{park2023choicemates,kaesberg2025voting}, and overall system capabilities. 

With the progress of multi-agent LLM systems, most existing frameworks primarily emphasize the overall design of agent collaboration workflows~\cite{tang2023medagents}. However, the scheduling of cooperative agents, which might heavily impact multi-agent collaboration effectiveness, remains an unexplored problem.
Recent empirical studies~\cite{dang2025multi,qian2024scaling} have demonstrated that the sequential arrangement of agent operations significantly influences systemic performance. 
Moreover, our experiments show that even when the same group of agents is involved, simply altering their execution sequence can lead to significant performance variations, as illustrated in Figure~\ref{fig:path_comparison}. 
Without \textbf{effective scheduling and routing mechanisms to dynamically coordinate agent behaviors}, multi-agent systems often suffer from  \textbf{inconsistent and suboptimal performance}, significantly hindering their applicability to complex real-world challenges.

To tackle this challenge, several methods have been proposed to improve the robustness and flexibility of agent scheduling. Early work relies on fixed pipeline workflows~\cite{qian2023chatdev}, where the execution process is rigid and lacks the capacity to accommodate different task needs. To introduce flexibility, some approaches hand over agent scheduling to LLMs~\cite{kim2024mdagents}. Nevertheless, the LLM's intrinsic uncertainty and opaque reasoning frequently yield inconsistent execution orders, eroding scheduling robustness. Recent approaches schedule agent cooperation with trainable interaction graphs~\cite{zhuge2024gptswarm,zhang2024g-designer}, yet these methods fix the graph topology at task definition or query input, preventing dynamic adjustments during inference and limiting adaptability to evolving task requirements.

In this study, we propose STRMAC, a state-aware routing framework for efficient multi-agent collaboration. It employs a lightweight language model to encode problem-solving states and leverages LLM-based embeddings (extracted via agent-specific prompts) to represent agent expertise. By dynamically selecting the most relevant agent at each step via cosine similarity matching between the problem-solving state and agent expertise, STRMAC enhances accuracy by 23.8\% while reducing inference costs and noise propagation. 
STRMAC learns to optimally select an agent at each collaboration step with a contrastive learning framework, minimizing redundant activations to reduce computational costs. However, router training becomes intractable due to factorial growth of execution paths as agent numbers increase, making exhaustive sampling prohibitively expensive. Consequently, an extremely large volume of path samples is required for training the router. To address the combinatorial path explosion, we introduce a self-evolving data generation strategy, cutting data collection overhead by 90.1\% compared to exhaustive search.

In summary, our contributions are as follows:

\begin{itemize}
    \item We propose STRMAC, a novel state-aware routing framework for multi-agent collaboration, which encodes evolving system states and leverages agent-specific expertise embeddings to enable targeted agent selection.
    \item We develop a self-evolving data generation pipeline that efficiently selects high-value agent paths and iteratively improves the router, greatly reducing the cost of training data collection.
    \item Experiments demonstrate that STRMAC has achieved substantial improvements in multi-agent reasoning accuracy, with gains of up to 23.8\% upon our datasets, while our data construction method reduces training data collection cost by up to 90.1\% compared to exhaustive search.
\end{itemize}

\section{Related Works}

\paragraph{Multi-Agent Collaboration}
Multi-agent collaboration with large language models involves combining multiple specialized agents to collectively address complex problems that are often beyond the reach of a single model. ChatDev~\cite{qian2023chatdev}, MedAgents~\cite{tang2023medagents}, and AgentVerse~\cite{chen2023agentverse}, introduce static, role-based pipelines, where agents with specialized roles operate according to predefined sequences. Recognizing the limitations of rigidity, subsequent research such as MDAgents~\cite{kim2024mdagents} and Beyond Frameworks~\cite{wang2025beyond} allows language models to manage agent coordination and task scheduling in a more dynamic and context-aware manner. Meanwhile, recent work—including GPTSwarm~\cite{zhuge2024gptswarm}, G-Designer~\cite{zhang2024g-designer}, DyLAN~\cite{liu2024dynamic}, and AgentPrune~\cite{zhang2024cut}—introduces trainable graph structures to model multi-agent communication as adaptive or optimizable graphs, enabling the collaboration structure and information flow to automatically adjust based on task requirements. 
Unlike dynamic agent selection mechanisms, these methods commit to specific interaction patterns before inference begins, resulting in (a) frozen interaction topology during runtime, and (b) diminishing adaptability as computational contexts change.

\paragraph{LLM Routing}
LLM routing seeks to efficiently allocate input to the most suitable model, maximizing quality while minimizing cost across diverse tasks. Methods such as FrugalGPT~\cite{chen2023frugalgpt}, Hybrid LLM~\cite{ding2024hybrid}, and RouteLLM~\cite{ong2024routellm} primarily focus on the two-model scenario, routing simple queries to weaker models and assigning complex or uncertain tasks to stronger models to balance cost and performance. ZOOTER~\cite{lu2023routing} extends routing beyond this binary setting by leveraging supervision from off-the-shelf reward models to enable expert selection among multiple LLMs with enhanced robustness. Capability Instruction Tuning~\cite{zhang2025capability} introduces model capability vectors to facilitate instruction-level alignment, allowing flexible adaptation to new tasks and models without requiring direct output queries. RouterDC~\cite{chen2024routerdc} further refines routing through contrastive learning, distinguishing fine-grained model capabilities and enabling efficient, generalizable, end-to-end allocation among multiple LLMs. Although these approaches advance the flexibility and efficiency of LLM routing, the predominant research focus remains constrained by static configurations and single-agent frameworks.
Building upon these foundational studies, the incorporation of LLM routing within a multi-agent architectural paradigm facilitates: (1) real-time adaptive collaboration, and (2) optimized task arrangement across multiple agents.

\section{Method}

\begin{figure*}[t]
\centering
\includegraphics[width=1.0\textwidth]{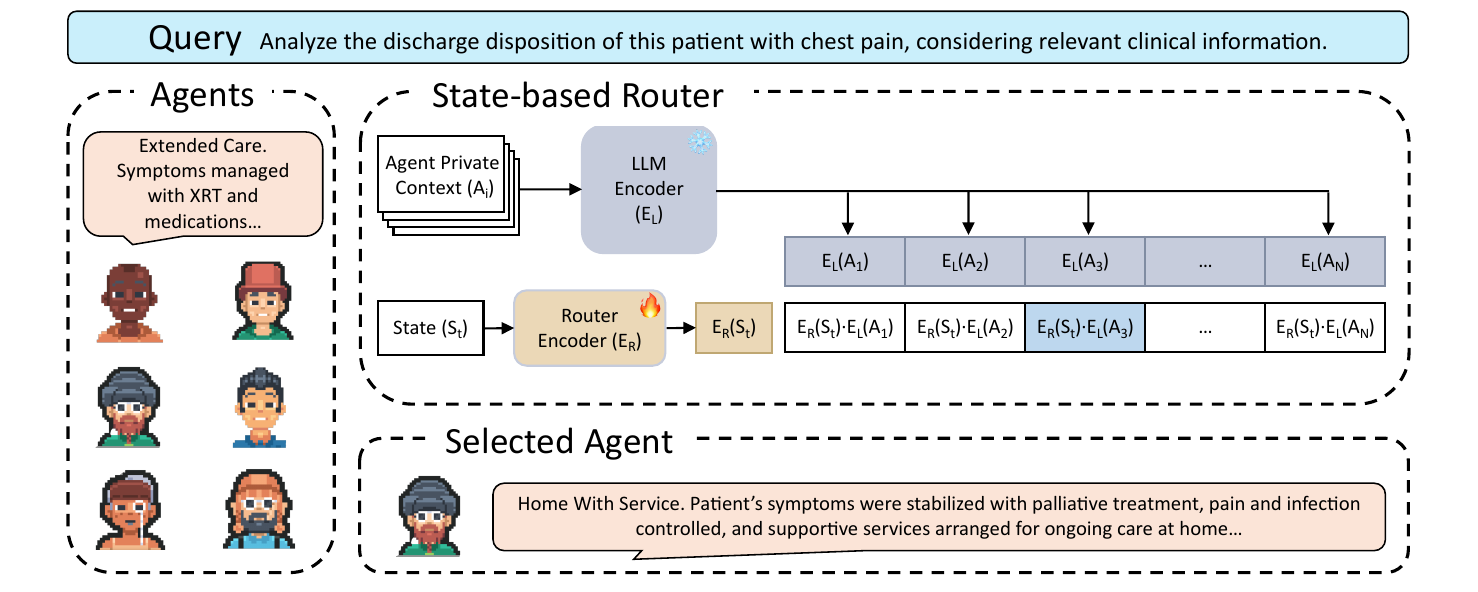} 
\caption{The overview of STRMAC framework. Each agent’s private context \(A_i\) is encoded by a LLM encoder to obtain \(E_L(A_i)\).
At each step \(t\), the current state \(s_t\) is encoded by the router to \(E_R(s_t)\), and the agent whose embedding best matches the state is selected to generate the output, updating the state for the next step.}
\label{overview}
\end{figure*}

In this section, we present our overall methodological framework for collaborative multi-agent decision making. Section 3.1 introduces the task formulation and agent interaction setting. Section 3.2 details the state-aware routing module for agent scheduling. Section 3.3 describes the contrastive learning objective for optimizing agent selection. Section 3.4 introduces a self-evolving data generation paradigm for efficient construction of high-quality training paths.

\subsection{Task Formulation}
In real-world scenarios, multiple experts with access to private information or unique expertise are able to independently analyze the entire problem and contribute their insights throughout a collaborative process. Inspired by this, we consider a general and flexible multi-agent setting, where a set of $N$ agents, denoted as ${A_i (i = 1, \ldots, N)}$, each have the ability to solve the complete task on their own. Instead of assigning agents to fixed roles or predetermined stages, our framework allows any agent to participate at any stage of the process, supporting dynamic collaboration and adaptive decision making.

The environment is represented by the system state $s$, which consists of the initial task query $q$ together with the accumulated history $h$ of agent actions and responses up to the current step. At each decision step, the router takes the current system state $s_t$ as input and outputs a probability distribution over the $N$ agents, indicating the suitability of each agent to act next. In this way, the system can adaptively select the most appropriate agent based on the evolving context and the history of the collaboration.

\subsection{State-Aware Router for Multi-Agent System}

Recent work such as RouterDC~\cite{chen2024routerdc} has shown that contrastive learning can substantially improve routing by leveraging the distinct strengths and specialized capabilities of different LLMs and enabling efficient, generalizable assignment across multiple models. Building on these insights, we design a state-aware router that makes routing decisions based on the evolving system state and the particular strengths of each agent. This allows the system to dynamically select the most suitable agent at each step of the collaborative process.

\paragraph{Router Encoder}

We implement the state encoder $E_R$ as a trainable lightweight language model. At each decision step, the encoder receives the current system state $s_t$, which comprises the initial task query $q$ and the multi-agent system history $h_t$. The encoder then transforms this information into a fixed-dimensional latent representation:
\begin{equation}
z_t = E_R(s_t)
\end{equation}
where $E_R(\cdot)$ denotes the router encoder model, which is jointly optimized during training. The resulting state embedding  $z_t \in R^d$ encodes the current system state and serves as the basis for subsequent routing decisions.

\paragraph{Agent Embedding}
To represent each agent’s unique expertise, we construct an agent embedding $e_i$ by encoding its private information or knowledge context using a large language model $E_L$. Specifically, for each agent $A_i$, we input its domain-specific knowledge or private data into the language model and use the resulting semantic embedding as $e_i$. These agent embeddings capture the distinct capabilities and information associated with each agent, allowing the router to identify and leverage the most suitable expertise during collaboration.

\paragraph{Routing Decision}
Given the state embedding $z_t$ and the set of agent embeddings ${e_i (i = 1, \ldots, N)}$, the router computes a compatibility score between the current state and each agent:
\begin{equation}
\mathrm{score}_i = \langle z_t, e_i \rangle
\end{equation}
where $\langle \cdot, \cdot \rangle$ denotes cosine similarity. The scores are then normalized with a softmax to produce a probability distribution over agents. At each decision step, the agent with the highest probability is selected to execute the next action. The system state is then updated with the agent's output, and the routing process iterates until task completion or a predefined termination criterion is met.

\subsection{Contrastive Learning for Router Training}

We adopt a contrastive learning objective to train the router encoder to accurately select the most appropriate agent for each task context. The core idea is to align the state embedding produced by the router encoder with the embedding of the agent best suited to the current context, while discouraging similarity with embeddings of other agents.

Formally, for each decision step $t$ in the training set, let $z_t$ denote the state embedding from the router encoder, and $e_{i^*}$ denote the embedding of the agent that achieves the optimal outcome for the given context. The contrastive loss is defined as
\begin{equation}
\mathcal{L}_{\text{contrast}} = -\log \frac{\exp(\langle z_t, e_{i^*} \rangle)}{\sum_{j=1}^N \exp(\langle z_t, e_j \rangle)}
\end{equation}
where $\langle \cdot, \cdot \rangle$ represents cosine similarity

During training, only the parameters of the router encoder are updated, while the agent embeddings generated in advance by the large language model remain fixed. This setup enables the router encoder to learn state representations that facilitate accurate identification of the optimal agent for each task context, thereby improving agent selection performance in multi-agent collaboration.

\subsection{Self-Evolving Data Generation}

A fundamental challenge in multi-agent routing is the efficient construction of high-quality training paths, given the factorial growth in the number of possible agent execution sequences. For each task with $N$ agents, where each agent participates at most once, all possible execution paths can be organized into a tree structure. In this tree, each node corresponds to a partial sequence of agent selections, and each path from the root to any node represents a valid execution path. The total number of candidate paths is given by the sum over all possible permutations of $K$ agents selected from $N$:
\begin{equation}
\mathrm{Path}(N) = \sum_{K=1}^{N} P(N, K) = \sum_{K=1}^{N} \frac{N!}{(N-K)!}
\label{eq:path_num}
\end{equation}
where $P(N, K)$ denotes the number of $K$-length permutations from $N$ agents. As $N$ increases, the execution path tree expands rapidly and soon becomes computationally intractable. To address this, we propose a self-evolving data generation framework that combines exhaustive search, principled pruning, and model-guided exploration to efficiently identify and collect high-value execution paths for router training.

\paragraph{Path Scoring}
Each candidate execution path $\mathcal{P}$ is scored using a clear criterion: only paths that yield the correct answer are considered, and among them, the path with the lowest token consumption is selected as optimal. Formally,
\begin{equation}
\mathrm{Score}(\mathcal{P}) =
\begin{cases}
    -\mathrm{TokenCount}(\mathcal{P}), & \text{if } \hat{y}^{(\mathcal{P})} = y \\
    -\infty, & \text{otherwise}
\end{cases}
\end{equation}
where $\mathrm{TokenCount}(\mathcal{P})$ denotes the total number of tokens consumed along path $\mathcal{P}$, $\hat{y}^{(\mathcal{P})}$ is the prediction corresponding to $\mathcal{P}$, and $y$ denotes the ground-truth answer. The score is assigned only if $\hat{y}^{(\mathcal{P})} = y$; otherwise, the score is $-\infty$.

\paragraph{Early Pruning with Solution-Aware Search}
The data generation process begins with an early-stopping pruning strategy inspired by Monte Carlo Tree Search (MCTS). Rather than exhaustively expanding every branch, the search terminates along each branch as soon as a correct solution is found, immediately discarding all longer descendant paths from that node. For each such branch, only the most efficient valid path is retained, reducing redundant computation and focusing search resources on unresolved parts of the tree.

In addition to the globally optimal path, alternative valid paths that reach correct answers are also collected across the search tree, even if they involve non-optimal agent selections at earlier steps. These alternative valid paths serve as auxiliary training signals and enable the router to learn how to recover from earlier non-optimal choices, ultimately still reaching correct solutions through subsequent decisions.

\paragraph{Router-Guided Iterative Data Construction}
The iterative data generation process begins by applying solution-aware pruning to obtain a small initial set of valid execution paths. This initial dataset is used to train a preliminary router that estimates agent selection probabilities at each decision step. In subsequent rounds, data generation framework integrates solution-aware pruning with router-guided expansion. At each node of the search tree, the router predicts the top-$k$ most promising agents for expansion. For each expanded path, the search continues only until a correct solution is found; all further descendant paths from that node are then pruned. This combination of router guidance and early stopping ensures efficient exploration while preserving diverse and high-quality candidate paths. As new informative paths are discovered, the router is incrementally retrained on the enlarged dataset, enabling its predictions to further improve the efficiency and coverage of future search rounds. Through this iterative and self-improving mechanism, the training set is progressively enriched, supporting effective multi-agent learning.
\section{Experiments}

\subsection{Experimental Setup}

\begin{table*}[htb]
\centering
\begin{tabular}{clccccccccc}
\hline
\multirow{2}{*}{Task} &
  \multirow{2}{*}{Method} &
  \multicolumn{3}{c}{Llama3.1-70B} &
  \multicolumn{3}{c}{Llama3.1-8B} &
  \multicolumn{3}{c}{Qwen2.5-32B} \\
 &
   &
  \multicolumn{1}{l}{Acc} &
  Token &
  CAS &
  \multicolumn{1}{l}{Acc} &
  Token &
  CAS &
  \multicolumn{1}{l}{Acc} &
  Token &
  CAS \\\hline
\multirow{7}{*}{PDDP} &
  Random-Chain &
  59.6 &
  7628.7 &
  \multicolumn{1}{l}{27.8} &
  45.1 &
  8338.3 &
  19.6 &
  53.6 &
  8137.6 &
  23.8 \\
 &
  LLM-Debate &
  53.6 &
  24330.6 &
  4.7 &
  41.9 &
  22974.0 &
  4.2 &
  47.9 &
  18341.9 &
  7.7 \\
 &
  MACNET &
  60.5 &
  \underline{2810.9} &
  \underline{45.7} &
  48.1 &
  \underline{3004.7} &
  \underline{35.6} &
  51.3 &
  \underline{2843.2} &
  \underline{38.6} \\
 &
  MOA &
  56.6 &
  13986.0 &
  14.0 &
  52.0 &
  14618.8 &
  12.1 &
  48.2 &
  13670.8 &
  12.3 \\
 &
  MDAgents &
  53.9 &
  56004.0 &
  0.2 &
  44.9 &
  64023.8 &
  0.1 &
  44.9 &
  64023.8 &
  0.1 \\
 &
  IG-MAS &
  \underline{62.4} &
  5785.7 &
  35.0 &
  \underline{52.2} &
  4158.6 &
  34.4 &
  \underline{55.2} &
  4755.5 &
  34.3 \\
 &
  \textbf{STRMAC (ours)} &
  \textbf{64.0} &
  \textbf{794.5} &
  \textbf{59.1} &
  \textbf{53.2} &
  \textbf{897.1} &
  \textbf{48.6} &
  \textbf{55.3} &
  \textbf{972.2} &
  \textbf{50.2} \\\hline
\multirow{7}{*}{EBFC} &
  Random-Chain &
  79.2 &
  2811.7 &
  \underline{59.8} &
  65.4 &
  2996.1 &
  48.5 &
  48.5 &
  3009.6 &
  35.9 \\
 &
  LLM-Debate &
  56.4 &
  12523.4 &
  16.1 &
  43.6 &
  11988.3 &
  13.1 &
  65.4 &
  6686.8 &
  33.5 \\
 &
  MACNET &
  47.5 &
  \underline{902.7} &
  43.4 &
  46.5 &
  \underline{960.0} &
  42.2 &
  47.5 &
  \underline{1124.3} &
  42.4 \\
 &
  MOA &
  \underline{83.2} &
  6004.4 &
  45.6 &
  73.3 &
  6427.2 &
  38.5 &
  51.5 &
  6169.9 &
  27.8 \\
 &
  MDAgents &
  82.2 &
  35165.7 &
  2.4 &
  71.3 &
  44514.3 &
  0.8 &
  \textbf{72.3} &
  27817.9 &
  4.5 \\
 &
  IG-MAS &
  82.2 &
  5777.0 &
  46.1 &
  \underline{79.2} &
  2371.4 &
  \underline{62.5} &
  58.4 &
  2303.3 &
  \underline{46.4} \\
 &
  \textbf{STRMAC (ours)} &
  \textbf{85.2} &
  \textbf{338.0} &
  \textbf{82.4} &
  \textbf{80.2} &
  \textbf{182.3} &
  \textbf{78.8} &
  \textbf{72.3} &
  \textbf{486.2} &
  \textbf{68.9}\\\hline
\end{tabular}
\caption{Comparison with baselines on multi-agent collaboration performance for PDDP and EBFC tasks. Best results are shown in \textbf{bold}; and second-best results are \underline{underlined}.}
\label{main result}
\end{table*}


\paragraph{Dataset}
Our experimental evaluation is conducted on challenging collaborative reasoning tasks that are intentionally designed to capture the challenges of multi-agent reasoning with incomplete information, where each agent receives a distinct subset of information. These tasks require effective communication and information integration among agents to reach accurate collective decisions. In particular, we use PDDP (Patient Discharge Disposition Prediction) and EBFC (Evidence-Based Fact-Checking), both constructed following the protocols of~\cite{wang2025beyond}, as our target tasks.

\textbf{PDDP (Patient Discharge Disposition Prediction)}
PDDP is a clinical prediction task constructed from the MIMIC-III database~\cite{johnson2016mimic}, where the objective is to infer a patient's discharge status based on fragmented medical records. Each agent accesses a distinct, non-overlapping subset of the patient's information as private input, while the chief complaint is provided as shared context. Through communication and integration of complementary information, agents collectively predict the patient's discharge disposition from a set of categories (expired, extended care, home with service, or home).

\textbf{EBFC (Evidence-Based Fact-Checking)}
EBFC is a fact verification task built on the AMBIFC dataset~\cite{glockner2024ambifc}, which contains claims paired with evidence sentences labeled as supporting, refuting, or neutral. In each instance, only one agent receives a key supporting or refuting evidence, while all other agents are assigned distractor evidence. Agents are required to engage in collaborative reasoning to identify relevant evidence, mitigate the influence of distractor information, and accurately determine the veracity of the claim.

\paragraph{Baselines}
We compare our approach with the following representative methods: (i) Random-Chain, which randomly assigns a sequential order to the agents, allowing each agent to process the task once in turn; (ii) LLM-Debate~\cite{du2023improving}, which improves solution quality via iterative proposal where each agent outputs depending only on previous round’s outputs of all agents.
(iii) MACNET~\cite{qian2024scaling}, a topology-driven agent collaboration framework based on directed acyclic graphs; (iv) MOA~\cite{wang2024mixture}, which uses layered aggregation and synthesis to combine multiple LLM outputs; (v) MDAgents~\cite{kim2024mdagents}, an adaptive multi-agent medical reasoning framework that allows agents to autonomously interact and persuade each other in order to reach a collective decision; and (vi) IG-MAS~\cite{wang2025beyond}, which implements the centralized, instructor-led governance, selective participation, ordered interaction, and context summarization strategy to optimize multi-agent collaboration and efficiency.

\paragraph{Implementation Details}
In our experiments, we use Llama-3.1-70B-Instruct~\cite{grattafiori2024llama}, Llama-3.1-8B-Instruct~\cite{grattafiori2024llama}, and Qwen2.5-32B-Instruct~\cite{hui2024qwen2} as agent models, all of which are open-source models available on HuggingFace. We also conduct supplementary experiments using the proprietary commercial model GPT to further validate the generality of our approach. Agent embeddings are obtained from the embedding layers of the respective large language models that each agent is based on. For the router encoder, we follow the configuration of RouterDC~\cite{chen2024routerdc} and employ mDeBERTaV3-base~\cite{he2020deberta} as the state encoder, a small language model with 86M parameters. The learning rate is tuned within the range of $1 \times 10^{-4}$ to $5 \times 10^{-5}$ according to validation performance. All experiments are conducted on A100 with 80G memory.

\subsection{Results}

\paragraph{Main Experimental Results}

We report the comparative results of STRMAC and mainstream baselines on both the PDDP and EBFC tasks in Table~\ref{main result}. All experiments are conducted under three different large language model, providing a comprehensive evaluation across various model capacities.

On the PDDP task, our method achieves state-of-the-art performance across all model configurations, reaching accuracies of 64.0\%, 53.2\%, and 55.3\%. Compared to the Random-Chain baseline, this corresponds to improvements of +4.4\%, +8.1\%, and +1.7\% under different settings. 
Our method dynamically selects the optimal agent for information integration based on real-time problem-solving needs. This approach provides superior capability in handling the fragmented and heterogeneous information characteristic of the PDDP task, yielding significant performance improvements over standard pipeline-based baselines.

On the EBFC task, our method delivers substantial accuracy gains, achieving 85.2\%, 80.2\% and 72.3\% under different model configurations, with absolute improvements of +6.0\%, +14.8\% and +23.8\% over the Random-Chain baseline. Notably, this scenario requires the framework to identify and amplify agents holding correct evidence while filtering out misleading or irrelevant information (i.e., noise) from other agents. In contrast, the MACNET approach, which constructs agent communication topologies based on random probabilities, struggles to effectively distinguish reliable sources from distractors, resulting in inferior performance. These results demonstrate the advantage of our framework in navigating complex, high-noise environments that require selective aggregation of reliable information from multiple sources.

\paragraph{Token Efficiency Analysis}

Inspired by the Constant Risk Aversion (CRA) utility framework~\cite{pratt1978risk}, we propose the Cost-adjusted Score (CAS) to effectively evaluate each method's trade-off between accuracy and inference cost. CAS penalizes accuracy based on token consumption and is defined as:
\begin{equation}
\text{CAS} = \text{Acc} \times \exp\left(-\mu \frac{\text{\#Token}}{c}\right),
\end{equation}
where $\mu$ is a cost sensitivity parameter and $c$ is a normalization constant. In our experiments, we set $\mu=0.1$ and $c=1000$, which means that every additional 1000 tokens penalizes the score by a factor of $e^{-0.1}$. By incorporating a cost adjustment inspired by CRA, CAS provides a practical measure of real-world effectiveness for large language model systems, especially in scenarios where inference cost matters.

From Table~\ref{main result}, STRMAC consistently achieves the highest CAS across all models and tasks. Compared to IG-MAS, STRMAC attains similar or higher accuracy while reducing token consumption to just 5.9\%–21.6\% of IG-MAS across different settings. For other baselines such as MDAgents, token usage reaches 57.2–244.2 times that of STRMAC, which results in CAS values less than 10\% of those achieved by STRMAC, despite comparable or even lower accuracy. 

\section{Analysis}

\begin{table}[t]
\centering
\setlength{\tabcolsep}{1mm}
\begin{tabular}{cccccc}
\hline
\multirow{2}{*}{Method} & \multicolumn{2}{c}{PDDP} & \multicolumn{2}{c}{EBFC} \\
                        & \#Path   & Proportion & \#Path   & Proportion  \\\hline
Llama3.1-70B            & 44.1     & 13.6       & 126.1    & 9.9        \\
Llama3.1-8B             & 51.7     & 15.9       & 141.8    & 11.1       \\
Qwen2.5-32B             & 45.3     & 13.9       & 170.9    & 13.4       \\
Exhaustive Search       & 325.0    & 100.0      & 1275.1   & 100.0      \\\hline
\end{tabular}
\caption{Performance of different methods under self-evolving data generation. \#Path: average number of sampled paths per instance. Proportion (\%): proportion of sampled paths relative to exhaustive search. Exhaustive Search: theoretical maximum from full enumeration.}
\label{self-evolving data generation}
\end{table}

\subsection{Effectiveness of Data Generation Strategy}

Table~\ref{self-evolving data generation} reports the weighted average number of sampled execution paths per instance on both the PDDP and EBFC datasets using our self-evolving data generation framework. In our experiments, 20\% of the training data is constructed using solution-aware pruning, and the remaining 80\% is generated by combining pruning with router-guided exploration.

The total number of execution paths per instance is determined by the sum of all possible agent permutations, as defined in Equation~(\ref{eq:path_num}). For PDDP, which involves five agents, this corresponds to 325.0 possible paths per instance. In EBFC, where the number of agents varies across examples, the average number of paths is 1275.1 per instance.

With our framework, the average number of sampled execution paths is reduced to just 13.6\%--15.9\% of the full permutation space in PDDP and 9.9\%--13.4\% in EBFC. This substantial compression of the search space enables much more efficient data construction and significantly lowers computational overhead. 

\paragraph{Iterative Router Improvement}
After initial training with data generated by solution-aware pruning, we further continue training the models with additional data collected through router-guided exploration. This incremental training with iteratively improved data yields up to 4.4\% higher accuracy on PDDP and 11.9\% on EBFC, demonstrating the progressive benefit of our self-evolving data generation pipeline.

\begin{table}[t]
\centering
\setlength{\tabcolsep}{1mm}
\begin{tabular}{cccccc}
\hline
\multirow{2}{*}{Agent Model} & \multirow{2}{*}{Training Source} & \multicolumn{2}{c}{PDDP} & \multicolumn{2}{c}{EBFC} \\
                          &            & \multicolumn{1}{l}{Acc} & Token  & \multicolumn{1}{l}{Acc} & Token \\\hline
\multirow{2}{*}{Llama70B} & Same & 64.0                    & 794.5  & 85.2                    & 338.0 \\
                          & Llama8B    & 63.4                    & 1166.6 & 83.2                    & 334.4 \\
\multirow{2}{*}{Qwen32B}  & Same & 55.3                    & 972.2  & 72.3                    & 486.2 \\
                          & Llama8B    & 55.5                    & 736.7  & 71.3                    & 439.6\\\hline
\end{tabular}
\caption{Performance of routers trained on data collected from different models. Agent Model: model used for inference in the multi-agent system. Training Source: model used to generate execution paths for router training. Same: data collected using the same model as Agent Model.}
\label{small router}
\end{table}

\subsection{Generalization Across Training Data Sources}

Table~\ref{small router} compares the performance of routers trained with data collected from different models in both the PDDP and EBFC scenarios. The results indicate that routers trained on data generated by the smaller Llama8B model achieve accuracy and token efficiency comparable to those trained with data from the same (larger) agent model. For example, with Llama70B as the agent model, accuracy differences between the two training sources are marginal for both PDDP (64.0\% vs.\ 63.4\%) and EBFC (85.2\% vs.\ 83.2\%), and token consumption remains similar. Comparable patterns are observed for Qwen32B.

These findings demonstrate strong generalization capability with respect to the training data source: a router trained on execution data generated by a smaller, more resource-efficient model can still deliver effective and efficient multi-agent coordination when deployed alongside larger language models. This substantially reduces the demand for large-model data collection and enhances the scalability and practicality of the overall framework.

\begin{table}[t]
\centering
\begin{tabular}{ccccc}
\hline
\multirow{2}{*}{Method} & \multicolumn{2}{c}{PDDP}       & \multicolumn{2}{c}{EBFC}       \\
                        & Acc           & Token          & Acc           & Token          \\\hline
Random-Chain            & 56.6          & 7934.3         & 76.2          & 2824.2         \\
LLM-Debate              & 53.6          & 23901.0        & 59.4          & 7056.5         \\
MACNET                  & 57.7          & 2826.4         & 64.4          & 1018.0         \\
MOA                     & 53.1          & 14236.6        & 43.6          & 6331.2         \\
MDAgents                & 54.6	        & 79759.8        & 81.2          & 45034.1        \\
IG-MAS                  & \underline{63.4}    & \underline{6322.9}   & \underline{83.2}    & \underline{4014.8}   \\
\textbf{STRMAC (ours)}    & \textbf{64.4} & \textbf{756.3} & \textbf{89.1} & \textbf{293.5}\\\hline
\end{tabular}
\caption{Comparison with baselines on GPT-4o for multi-agent collaboration performance on PDDP and EBFC tasks. Best results are in bold; second-best results are underlined.}
\label{gpt4o_results}
\end{table}

\subsection{Performance on  GPT-4o}

To further evaluate the generalization capability of the proposed STRMAC method on commercial closed-source models, we selected OpenAI's GPT-4o~\cite{OpenAI2024GPT-4o} as the agent model for experiments. Considering the high cost of inference and data collection on GPT-4o, instead of constructing a full-scale trajectory dataset on this model, we leveraged multi-agent execution data collected from previous open-source models (Llama3.1-70B-Instruct, Llama3.1-8B-Instruct, Qwen2.5-32B-Instruct), filtering out high-quality and high-value execution paths to form the training set for the GPT-4o router. Additionally, as GPT-4o does not provide access to its embedding interface, we utilized agent embeddings derived from Llama3.1-70B-Instruct as a substitute.

The experimental results, as shown in Table~\ref{gpt4o_results}, demonstrate that, even with training data and embeddings sourced from open models, STRMAC achieves state-of-the-art multi-agent collaboration performance on GPT-4o. Specifically, on the PDDP and EBFC tasks, STRMAC achieves accuracies of 64.4\% and 89.1\% respectively, significantly outperforming mainstream baselines. Moreover, STRMAC dramatically reduces token consumption---requiring only 756.3 tokens for PDDP and 293.5 tokens for EBFC---substantially less than competing multi-agent coordination approaches. These results strongly validate the efficiency and transferability of our approach in high-cost, closed-source scenarios.

\subsection{Path Selection Analysis}
This section examines the routing behaviors and path selection distributions of the multi-agent system on the PDDP task. Prior to analyzing multi-agent path selection, we first assess the independent reasoning performance of each individual agent. As Table~\ref{individual agent}, the agent provided with the Brief Hospital Course (\(Agent_{\text{BHC}}\))  attains the highest accuracy across all three models, owing to its comprehensive overview of the patient's clinical trajectory. In contrast, the other agents, with limited and fragmented information, exhibit inferior performance.

\begin{table}[ht]
\centering
\begin{tabular}{cccc}
\hline
Methods   & \multicolumn{1}{l}{Llama70B} & Llama8B & Qwen32B \\\hline
$Agent_{\text{BHC}}$  & 62.4                         & 51.0    & 56.1    \\
$Agent_{\text{DM}}$   & 49.2                         & 39.8    & 41.3    \\
$Agent_{\text{MSIP}}$ & 49.6                         & 44.4    & 41.8    \\
$Agent_{\text{SH}}$   & 44.4                         & 31.0    & 39.0    \\
$Agent_{\text{PR}}$    & 44.6                         & 44.2    & 41.1    \\\hline
\end{tabular}
\caption{Comparison of individual agent performance for the PDDP task. BHC: Brief Hospital Course. DM: Discharge Medications. MSIP: Major Surgical or Invasive Procedure. SH: Social History. PR: Pertinent Results.}
\label{individual agent}
\end{table}

\begin{figure}[t]
\centering
\includegraphics[width=1.0\columnwidth]{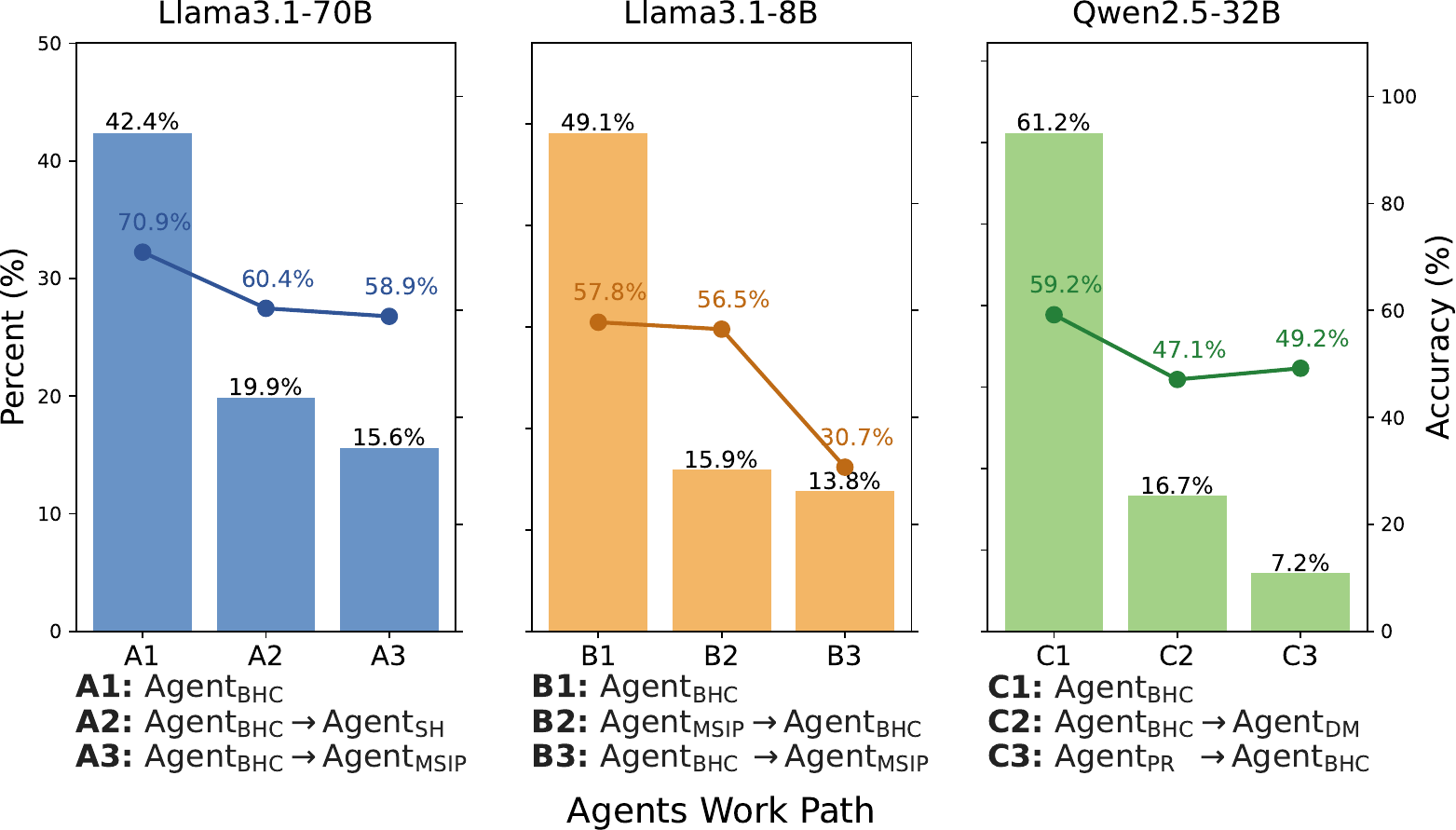} 
\caption{Distribution (bar, left axis) and accuracy (line, right axis) of the top-3 most frequently selected agent work paths for each model in the PDDP task.}
\label{fig:path}
\end{figure}

Figure~\ref{fig:path} shows the distribution and accuracy of the most frequently selected routing paths. Across all three models, the path invoking only \(Agent_{\text{BHC}}\) overwhelmingly dominates, indicating the prioritization of high-value information. However, in some cases, collaborative paths such as \(Agent_{\text{BHC}}\) $\rightarrow$ \(Agent_{\text{SH}}\) or \(Agent_{\text{BHC}}\) $\rightarrow$ \(Agent_{\text{MSIP}}\) are also selected, suggesting that supplementary information from other agents can sometimes enhance prediction. Notably, these agent collaborative paths do not always result in higher accuracy, suggesting that only carefully selected, relevant agents can provide meaningful improvements, while the inclusion of less relevant agents may introduce noise and negatively impact performance.

Interestingly, the observed path selection behaviors align well with clinical reasoning patterns in real-world practice: clinicians typically rely on the most comprehensive and relevant information source for decision-making and consult additional sources only when necessary. This similarity highlights the interpretability and rationality of the routing framework.
\section{Conclusion}
In this study, we propose STRMAC, a state-aware routing framework that enables efficient and targeted collaboration among multiple LLM agents. Through state encoding, agent-specific embeddings, and a contrastive learning objective, STRMAC ensures effective agent selection and minimizes inference cost. An adaptive data generation strategy further enhances training efficiency by focusing on valuable agent trajectories. Experimental results confirm that STRMAC significantly improves multi-agent reasoning performance while greatly reducing data collection overhead, establishing it as an effective and intelligent solution for multi-agent collaboration.

\bibliography{aaai2026}
\end{document}